\documentclass[sigconf]{acmart}

\usepackage{booktabs} 
\usepackage{bm}
\usepackage{tikz}
\usepackage{etoolbox,siunitx}
\usepackage{multirow}
\usetikzlibrary{calc,decorations.markings,math}
\usetikzlibrary{matrix,chains,positioning,arrows,shapes,decorations.pathreplacing,arrows.meta,shapes.multipart}
\usepackage{todonotes}

\setcopyright{rightsretained}

\renewcommand\footnotetextcopyrightpermission[1]{} 
\begin{document}
\title[TransNet: A deep network for fast detection of common shot transitions]{TransNet: A deep network for fast detection of common shot transitions}

\author{Tom\'{a}\v{s} Sou\v{c}ek, Jaroslav Moravec, Jakub Loko\v{c}}
\affiliation{%
  \institution{SIRET Research Group, Department of Software Engineering \\
Faculty of Mathematics and Physics, Charles University, Prague, Czech Republic}
}
\email{tomas.soucek1@gmail.com, lokoc@ksi.mff.cuni.cz}


\begin{abstract}
Shot boundary detection (SBD) is an important first step in many video processing applications. This paper presents a simple modular convolutional neural network architecture that achieves state-of-the-art results on the RAI dataset with well above real-time inference speed even on a single mediocre GPU. The network employs dilated convolutions and operates just on small resized frames. The training process employed randomly generated transitions using selected shots from the TRECVID IACC.3 dataset. The code and a selected trained network will be available at \url{https://github.com/soCzech/TransNet}.
\end{abstract}

%
%

\settopmatter{printfolios=true}

\maketitle

\section{Introduction}
A popular way to structure a video is by making use of a shot composition, where shots are delimited by transitions. Since information about the transitions is not available in the video format, automated shot boundary detection is an important step for video management and retrieval systems. For example, information about shots can be employed for video summarization, advanced browsing and filtering in known-item search tasks \cite{CobarzanSBHBLVB17,LokocBSMA18,Lokoc2019}.
Shot changes can be either immediate (hard cuts) or gradual, the later spanning from basic linear interleaving of two shots over a certain number of video frames to more exotic geometric transformations from one shot to another one.
To make matters worse shot boundary detectors must distinguish between shot transitions and sudden changes in a video caused by partial occlusion of the scene by an object passing closer to the camera. Fast camera motion or motion of an object in the scene also should not be mistaken for a shot transition. This may indicate that some semantic representation of a scene is necessary to correctly segment a video.

In this work, we propose TransNet, a scalable architecture with multiple dilated 3D convolutional operations per layer (instead of only one as is usual) resulting in the greater field of view with less trainable parameters. Even though the architecture is trained on just two common types of transitions (hard cuts and dissolves), it achieves state-of-the-art results on the RAI dataset \cite{Baraldi15}.

\section{Related work}
The goal of the shot boundary detection is to temporally segment a video into shots. To determine the shot boundary, one of the first methods utilized thresholded pixel differences \cite{zhang93} effective for stationary shots with a small number of moving objects. Since then, more robust techniques to compare images were developed based on local color histograms, color coherence vectors \cite{Pass1997} or SIFT features. The work of Shao et al. \cite{shao15} utilizes HSV and gradient histograms for shot boundary detection, Apostolidis et al. \cite{apostolidis14} use not only the histogram but also a set of SURF descriptors to detect the differences between a pair of frames. Other approaches revolve around edge information \cite{Huan2008} or motion vectors \cite{amel10}. 

With the advent of deep learning, new methods for shot detection using convolutional neural networks (CNN) emerged. Baraldi et al. \cite{Baraldi15} utilize spectral clustering given a set of features for every frame extracted by a deep siamese network. Recently, Gygli \cite{Gygli18} used a relatively shallow neural network with 3D convolutions with the third dimension over time. Even though 3D convolutions significantly increase computational complexity and memory requirements over standard 2D convolutions due to the added dimension, Gygli has beaten the previous approach in accuracy and speed as well. Another approach by Hassanien et al. \cite{Hassanien17} also uses 3D CNN however its output is fed through SVM classifier and further postprocessing is done to reduce false alarms of gradual transitions through a histogram-driven temporal differencing.
Our work partially overcomes problem of computationally hungry 3D convolutions when a large field of view is required to cope with long gradual transitions by using dilated convolutions over the time dimension, which had been proven useful in speech generation task \cite{oord16}.

The deep learning approaches revolve around the need for large annotated datasets. Until recently \cite{Tang2018}, the size of publicly available datasets for SBD was the limiting factor. Fortunately, synthetic training data can be easily generated from virtually any video content by interleaving randomly selected sequences from different videos as is done in \cite{Gygli18} and others. The downside of this method is, however, that the real data can contain cuts between shots of the same scene which rarely occur in the synthetic data sets due to the nature how they are generated.

\section{Model architecture}

\begin{figure}[t]
    \centering
    \begin{tikzpicture}[
    box/.style={
    	draw,
    	minimum width=0.7cm,
    	minimum height=0.4cm,
    	font=\scriptsize,
    	rounded corners=2, inner sep=2pt, align=center, anchor=north,
    	text width=1.7cm
    }, pil/.style={
    	-{Stealth[scale=.5]},
    	rounded corners=5pt,
    	line width=1pt
    }]
    \definecolor{my-green}{RGB}{208, 240, 192}
    \definecolor{my-yellow}{RGB}{247, 231, 206}
    \definecolor{my-blue}{RGB}{195, 236, 255}
    \definecolor{my-gray}{RGB}{251, 251, 251}
    \definecolor{my-gray2}{RGB}{245, 245, 245}

    \node[box, draw=none, text width=3cm] (input) at (0,3.2) {Input\\$N \times width \times height \times 3$};
    
    \node[box, draw=lightgray, text width=8.2cm, text height=2.6cm, fill=my-gray] at (0,2.6) {};

    \node[box, draw=lightgray, text width=8cm, text height=1.6cm, fill=my-gray2] at (0,2.5) {};
    
    \node[box, fill=my-green] (layer1-1) at (-3,2) {Conv 3$\times$3$\times$3\\[-2pt]{\tiny dilation 1}};
    \node[box, fill=my-green] (layer1-2) at (-1,2) {Conv 3$\times$3$\times$3\\[-2pt]{\tiny dilation 2}};
    \node[box, fill=my-green] (layer1-3) at (1,2) {Conv 3$\times$3$\times$3\\[-2pt]{\tiny dilation 4}};
    \node[box, fill=my-green] (layer1-4) at (3,2) {Conv 3$\times$3$\times$3\\[-2pt]{\tiny dilation 8}};
    
    \node[box, text width=1.2cm] (layer1-out) at (0,1.3) {Concat};
    
    \node[anchor=south west] at (-4.1,.7) {\tiny \textit{DDCNN cell, each conv with $2^{i-1}F$ channels}};
    \node[anchor=south east, align=right] at (4.1,.7) {\scriptsize stack S times};
    
    \node[box, anchor=north, fill=my-yellow] (max-pool) at (0,0.6) {Max pooling\\1$\times$2$\times$2};
    
    \node[anchor=south west] at (-4.2,-0.2) {\tiny \textit{SDDCNN block}};
    \node[anchor=south east, align=right] at (4.2,-0.2) {\scriptsize stack L times};

    \node[box, anchor=north, fill=my-blue] (dense1) at (0,-.3) {Dense D};
    \node[box, anchor=north, fill=my-blue] (dense2) at (0,-.9) {Dense 2};
    \node[box, anchor=north, text width=1.2cm] (softmax) at (0,-1.5) {Softmax};
    \node[box, draw=none] (x) at (0,-2.1) {$N \times 2$};
    
    \draw[rounded corners=5pt, line width=1pt] (input) -- (0,2.3);
    
    \draw[pil] (0,2.3) -| (layer1-1);
    \draw[pil] (0,2.3) -| (layer1-2);
    \draw[pil] (0,2.3) -| (layer1-3);
    \draw[pil] (0,2.3) -| (layer1-4);
    
    \draw[pil] (layer1-1) |- (layer1-out);
    \draw[pil] (layer1-2) |- (layer1-out);
    \draw[pil] (layer1-3) |- (layer1-out);
    \draw[pil] (layer1-4) |- (layer1-out);
    
    \draw[pil] (layer1-out) -- (max-pool);
    \draw[pil] (max-pool) -- (dense1);
    \draw[pil] (dense1) -- (dense2);
    \draw[pil] (dense2) -- (softmax);
    \draw[pil] (softmax) -- (x);
    
    \end{tikzpicture}
    \caption{TransNet shot boundary detection network architecture for $S=1$ and $L=1$. \textmd{Note that $N$ represents length of video sequence, not batch size. In our case $N=100$.}}
    \label{fig:nn_architecture}
\end{figure}
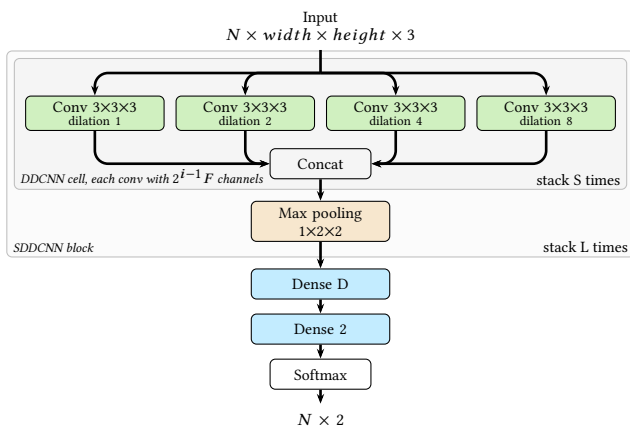

The proposed TransNet architecture (Figure \ref{fig:nn_architecture}) follows the work of Gygli \cite{Gygli18} and other standard convolutional architectures. As an input, the network takes a sequence of $N$ consecutive video frames and applies series of 3D convolutions returning a prediction for every frame in the input. Each prediction expresses how likely a given frame is a shot boundary.

The main building block of the model (Dilated DCNN cell) is designed as four 3D 3$\times$3$\times$3 convolutional operations. The convolutions employ different dilation rates for the time dimension and their outputs are concatenated in the channel dimension. This approach significantly reduces the number of trainable parameters compared to standard 3D convolutions with the same field of view. Multiple DDCNN cells on top of each other followed by spatial max pooling form a Stacked DDCNN block. The TransNet consists of multiple SDDCNN blocks, every next block operating on smaller spatial resolution but a greater channel dimension, further increasing the expressive power and the receptive field of the network.

Two fully connected layers refine the features extracted by the convolutional layers and predict a possible shot boundary for every frame representation independently (layers' weights are shared). ReLU activation function is used in all layers with the only exception of the last fully connected layer with softmax output. Stride 1 and the `same' padding is employed in all convolutional layers.

\section{Training}
This section describes the employed dataset and training settings.

\subsection{Dataset}
The TRECVID IACC.3 dataset \cite{2017trecvidawad} was utilized as it is provided with a set of predefined temporal segments. Hence, pairs of the predefined segments can be randomly selected from the pool for automatic creation of transitions for training purposes. More specifically, we considered segments of 3000 IACC.3 randomly selected videos. Furthermore, segments with less than 5 frames were excluded and from the remaining set only every other segment was picked, resulting in selected 54884 segments.

The training examples were generated on demand during training by randomly sampling two shots and joining them by a random type of a transition. Only hard cuts and dissolves were considered for training. Position of the transition was generated randomly. For dissolves, also its length was generated randomly from the interval $[5, 30]$. The length $N$ of each training sequence was selected to be 100 frames. The size of the input frames was set to $48\times 27$ pixels.

In order to validate the models, additional 100 IACC.3 videos (i.e., different from the training set) were manually labeled, resulting in 3800 shots. For testing, the RAI dataset \cite{Baraldi15} was considered.

\subsection{Training details}
The proposed architecture provides the following meta-parameters that were investigated by a grid search:

\begin{enumerate}
\item $S$, the number of DDCNN cells in a SDDCNN layer,
\item $L$, the number of SDDCNN layers,
\item $F$, the number of filters in the first set of DDCNN layers (doubled in each following SDDCNN layer),
\item $D$, the number of neurons in the dense layer.
\end{enumerate}

For training, batch size of 20 was used for all investigated networks. In order to prevent overfitting, only 30 epochs were  considered, each with 300 batches. Adam optimizer \cite{Adam14} with the default learning rate $0.001$ and cross entropy loss function were used. According to our preliminary evaluations, dropout did not improve results. Nevertheless, we plan to investigate advanced forms of regularization and training data augmentation in the future. Depending on the architecture, the whole training took approximately two to four hours to complete on one Tesla V100 GPU.

Even in the case of dissolves, when the transition is over multiple frames, the network was trained to predict only the middle frame as a shot boundary. This creates a discrepancy between the number of `transition' frames (each sequence contains only one) and frames without a transition (99 in our case). 
Increasing the weight of the transitions in the loss function did not produce better results than lowering the acceptance threshold $\theta$ under commonly used $0.5$; therefore, the latter approach is used.

\section{Evaluation}
During validation and testing, the list of shots is constructed in the following way: The shot starts at the first frame when the prediction drops below a threshold $\theta$ and ends at the first frame when the prediction exceeds $\theta$. The evaluation metric described in Section \ref{sec:eval_met} compares the generated shot list with the ground truth. Note that only predictions for frames 25-75 are used due to incomplete temporal information for the first/last frames. Therefore, when processing a video, the input window is shifted by 50 frames between individual forward passes through the network.

\subsection{Evaluation metric}
\label{sec:eval_met}
\begin{figure}[ht]
    \centering
    \includegraphics[width=.45\textwidth]{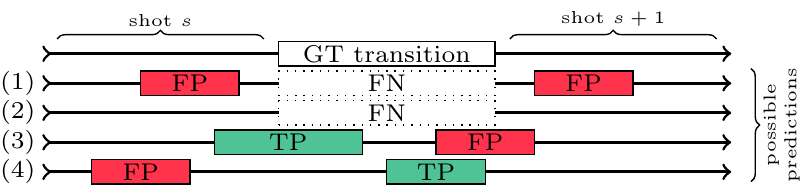}
    \caption{Visualization of the evaluation approach. \textmd{Predicted transitions shown with solid and missed with dotted rectangles.}}
    \label{fig:metricVis}
\end{figure}

The F1 score is used as an evaluation metric which is the same metric as in \cite{Baraldi15}. Reported F1 score is computed as an average of individual F1 scores for each video. Based on our analysis of the evaluation script\footnote{Source code of the evaluation method is available at \url{http://imagelab.ing.unimore.it/imagelab/researchActivity.asp?idActivity=19}}, Figure \ref{fig:metricVis} shows cases when detected shots are considered to be true positive, false positive, or false negative. A true positive is detected only if the detected shot transition overlaps with the ground truth transition (3, 4 in green). A false positive is detected if the predicted transition has no overlap with the ground truth (1, 4 in red) or the transition is detected for the second time (3 in red). A false negative is detected if there is no transition overlapping with the ground truth (1, 2 dotted) -- the ground truth transition is missed.

\subsection{Results}
\begin{figure}
    \centering
    \includegraphics[width=.46\textwidth]{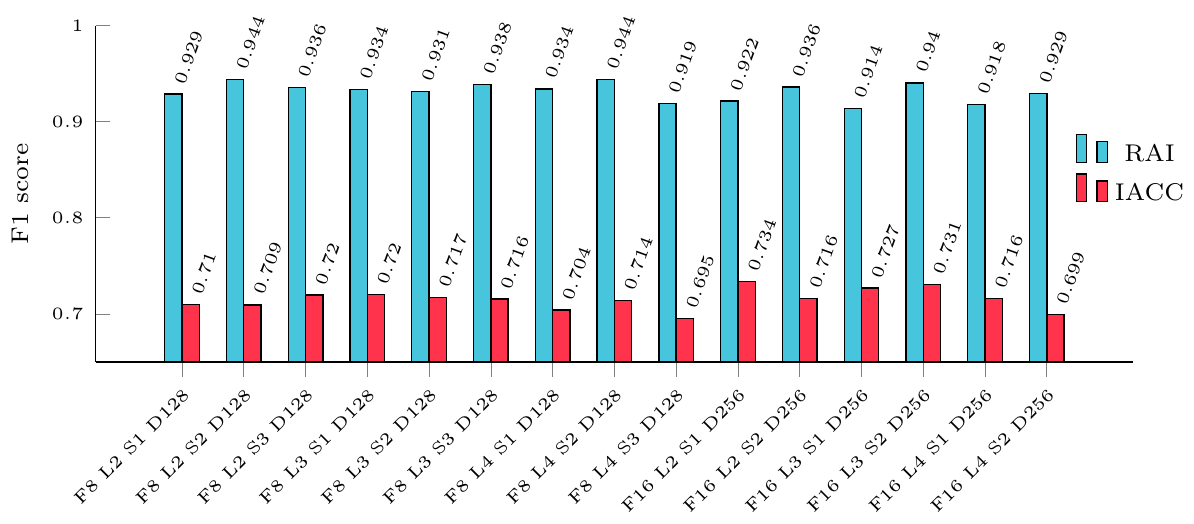}
    \caption{Observed average F1 scores of tested networks for the validation and test datasets.}
    \label{fig:F1scores}
\end{figure}

\begin{figure}
    \centering
    \includegraphics[width=.48\textwidth]{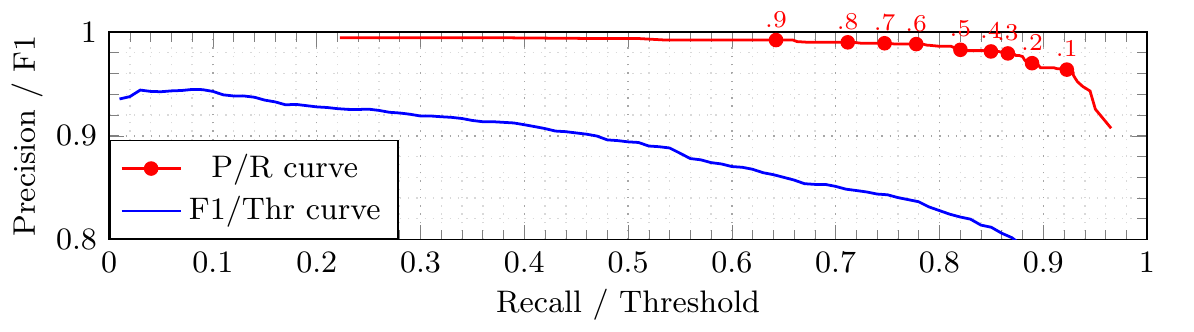}
    \caption{Precision/Recall curve for the best performing model with corresponding thresholds $\theta$ next to the points (in red) and F1 score dependency on threshold (in blue). Measured on RAI dataset.}
    \label{fig:prcurve}
\end{figure}

Figure \ref{fig:F1scores} presents the F1 scores of investigated models for validation and test datasets. Note that the top performing weights for each model configuration were selected based on results on validation dataset after each epoch. The confidence threshold $\theta$ indicating transition was set to $\theta=0.1$ as it performed reasonably well for most of the models. The effect of $\theta$ on precision, recall and F1 score is depicted in Figure \ref{fig:prcurve}.

Based on the evaluations presented in Figure \ref{fig:F1scores}, the best performing model is considered the one with 16 filters in the first layer, two stacked DDCNN cells in every one of the three SDDCNN blocks and with 256 neurons in the dense layer (F=16, L=3 S=2, D=256).
The average F1 score $0.94$ of the top performing model on the RAI dataset (see Table \ref{tab:shotDetectors}) is on par with the score reported by Hassanien et al. \cite{Hassanien17}. The overall F1 score even slightly outperforms the work of Hassanien et al., even though they proposed a network with more than 40 times as many parameters trained for a larger set of transition types. Furthermore, our model has the advantage that no additional post-processing is needed.

\begin{table}[b]
    \centering
    \begin{tabular}{r|c|c|c|c}
        & Baraldi et al. & Gygli & Hassanien et al. & ours \\
        \hline
        average & 0.84 \cite{Baraldi15} & 0.88 \cite{Gygli18} & $\mathbf{0.94}$ \cite{Hassanien17} & $\mathbf{0.94}$ \\
        overall & - & - & 0.934 \cite{Hassanien17} & $\mathbf{0.943}$ \\
    \end{tabular}
    
    \caption{Average and overall F1 scores for the RAI test dataset of the best architectures. \textmd{The overall F1 scores are computed by calculating precesion and recall over the whole dataset, not just single video.
    }}
    \label{tab:shotDetectors}
\end{table}

\begin{table}[b]
    \centering
    \begin{tabular}{|c|c|c|c|c|c|c|c|}
        \hline
        \textbf{Video} & \textbf{\#T}   & \textbf{TP}    & \textbf{FP}    & \textbf{FN}    & \textbf{P}     & \textbf{R}     & \textbf{F1}   \\
        \hline
        V1       &    80 &    57 &     2 &    23 & 0.966 & 0.713 & 0.820\\
        V2       &   146 &   132 &     5 &    14 & 0.964 & 0.904 & 0.933\\
        V3       &   112 &   111 &     4 &     1 & 0.965 & 0.991 & 0.978\\
        V4       &    60 &    59 &     5 &     1 & 0.922 & 0.983 & 0.952\\
        V5       &   104 &   101 &     8 &     3 & 0.927 & 0.971 & 0.948\\
        V6       &    54 &    53 &     3 &     1 & 0.946 & 0.981 & 0.964\\
        V7       &   109 &   103 &     1 &     6 & 0.990 & 0.945 & 0.967\\
        V8       &   196 &   181 &     4 &    15 & 0.978 & 0.923 & 0.950\\
        V9       &    61 &    55 &     2 &     6 & 0.965 & 0.902 & 0.932\\
        V10      &    63 &    57 &     0 &     6 & 1.000 & 0.905 & 0.950\\
        \hline
        Overall  &   985 &   909 &    34 &    76 & 0.964 & 0.923 & 0.943\\
        \hline
    \end{tabular}
    
    \caption{Per video results on the RAI dataset. \textmd{For each video the total number of transitions (\#T), true positives (TP), false positives (FP), false negatives (FN), precision (P), recall (R) and F1 score (F1) are shown.}}
    \label{tab:resultsRAI}
\end{table}

Since the validation dataset contains various sequences of frames where even annotators are not sure whether there is a shot transition, the reported scores for the validation data are lower. In addition, even the top performing TransNet model faces problems with detection of some transitions, for example, false positives in dynamic shots and false negatives in gradual transitions.

The model detected 1058 false positives and 679 false negatives with respect to the annotation. After closer inspection, for about 20\% of false negatives there was one very close false positive (shifted by one frame). This is in contrast to the RAI dataset results (Table \ref{tab:resultsRAI}) where the network achieves a lower number of false positives than false negatives. Based on manual inspection of the videos we conclude that RAI videos do not contain many highly dynamic shots (i.e. resulting in false positives) compared to the IACC.3 validation set.

\section{Conclusion}

In this paper, we present the TransNet neural network, the first shot detection model based on dilated 3D convolutions. The effectiveness of dilated 3D convolutions has been shown on RAI dataset with the TransNet performing on par with the current state-of-the-art approach without any additional post-processing and with a fraction of learnable parameters. The network also runs more than 100x faster than real-time on a single powerful GPU\footnote{It took just 50s to detect shot boundaries of preprocessed frames from the whole RAI dataset (about 98 minutes of video) using Tesla V100 GPU.}.

In the future, we plan to do further evaluation and improvements to enable deeper and more robust models.
The source code and our trained model will be available at \url{https://github.com/soCzech/TransNet}.

\begin{acks}
 This paper has been supported by Czech Science Foundation (GA\v{C}R) project Nr. 19-22071Y.
\end{acks}

\bibliographystyle{ACM-Reference-Format}
\bibliography{acmart}


\begin{thebibliography}{16}


\ifx \showCODEN    \undefined \def \showCODEN     #1{\unskip}     \fi
\ifx \showDOI      \undefined \def \showDOI       #1{#1}\fi
\ifx \showISBNx    \undefined \def \showISBNx     #1{\unskip}     \fi
\ifx \showISBNxiii \undefined \def \showISBNxiii  #1{\unskip}     \fi
\ifx \showISSN     \undefined \def \showISSN      #1{\unskip}     \fi
\ifx \showLCCN     \undefined \def \showLCCN      #1{\unskip}     \fi
\ifx \shownote     \undefined \def \shownote      #1{#1}          \fi
\ifx \showarticletitle \undefined \def \showarticletitle #1{#1}   \fi
\ifx \showURL      \undefined \def \showURL       {\relax}        \fi
\providecommand\bibfield[2]{#2}
\providecommand\bibinfo[2]{#2}
\providecommand\natexlab[1]{#1}
\providecommand\showeprint[2][]{arXiv:#2}

\bibitem[\protect\citeauthoryear{Amel, Abdelali, and Mtibaa}{Amel
  et~al\mbox{.}}{2010}]%
        {amel10}
\bibfield{author}{\bibinfo{person}{Abdelati~Malek Amel},
  \bibinfo{person}{Abdessalem~Ben Abdelali}, {and} \bibinfo{person}{Abdellatif
  Mtibaa}.} \bibinfo{year}{2010}\natexlab{}.
\newblock \showarticletitle{Video shot boundary detection using motion activity
  descriptor}.
\newblock \bibinfo{journal}{\emph{CoRR}}  \bibinfo{volume}{abs/1004.4605}
  (\bibinfo{year}{2010}).
\newblock
\showeprint[arxiv]{1004.4605}
\urldef\tempurl%
\url{http://arxiv.org/abs/1004.4605}
\showURL{%
\tempurl}


\bibitem[\protect\citeauthoryear{Apostolidis and Mezaris}{Apostolidis and
  Mezaris}{2014}]%
        {apostolidis14}
\bibfield{author}{\bibinfo{person}{Evlampios~E. Apostolidis} {and}
  \bibinfo{person}{Vasileios Mezaris}.} \bibinfo{year}{2014}\natexlab{}.
\newblock \showarticletitle{Fast shot segmentation combining global and local
  visual descriptors}.
\newblock \bibinfo{journal}{\emph{2014 IEEE International Conference on
  Acoustics, Speech and Signal Processing (ICASSP)}} (\bibinfo{year}{2014}),
  \bibinfo{pages}{6583--6587}.
\newblock


\bibitem[\protect\citeauthoryear{Awad, Butt, Fiscus, Michel, Joy, Kraaij,
  Smeaton, Qu\'{e}not, Eskevich, Ordelman, Jones, and Huet}{Awad
  et~al\mbox{.}}{2017}]%
        {2017trecvidawad}
\bibfield{author}{\bibinfo{person}{George Awad}, \bibinfo{person}{Asad Butt},
  \bibinfo{person}{Jonathan Fiscus}, \bibinfo{person}{Martial Michel},
  \bibinfo{person}{David Joy}, \bibinfo{person}{Wessel Kraaij},
  \bibinfo{person}{Alan~F. Smeaton}, \bibinfo{person}{Georges Qu\'{e}not},
  \bibinfo{person}{Maria Eskevich}, \bibinfo{person}{Roeland Ordelman},
  \bibinfo{person}{Gareth J.~F. Jones}, {and} \bibinfo{person}{Benoit Huet}.}
  \bibinfo{year}{2017}\natexlab{}.
\newblock \showarticletitle{TRECVID 2017: Evaluating Ad-hoc and Instance Video
  Search, Events Detection, Video Captioning and Hyperlinking}. In
  \bibinfo{booktitle}{\emph{Proceedings of TRECVID 2017}}. NIST, USA.
\newblock


\bibitem[\protect\citeauthoryear{Baraldi, Grana, and Cucchiara}{Baraldi
  et~al\mbox{.}}{2015}]%
        {Baraldi15}
\bibfield{author}{\bibinfo{person}{Lorenzo Baraldi},
  \bibinfo{person}{Costantino Grana}, {and} \bibinfo{person}{Rita Cucchiara}.}
  \bibinfo{year}{2015}\natexlab{}.
\newblock \showarticletitle{Shot and Scene Detection via Hierarchical
  Clustering for Re-using Broadcast Video}. In
  \bibinfo{booktitle}{\emph{Computer Analysis of Images and Patterns}},
  \bibfield{editor}{\bibinfo{person}{George Azzopardi} {and}
  \bibinfo{person}{Nicolai Petkov}} (Eds.). \bibinfo{publisher}{Springer
  International Publishing}, \bibinfo{address}{Cham},
  \bibinfo{pages}{801--811}.
\newblock
\showISBNx{978-3-319-23192-1}


\bibitem[\protect\citeauthoryear{Cob{\^{a}}rzan, Schoeffmann, Bailer,
  H{\"{u}}rst, Bla\v{z}ek, Loko\v{c}, Vrochidis, Barthel, and
  Rossetto}{Cob{\^{a}}rzan et~al\mbox{.}}{2017}]%
        {CobarzanSBHBLVB17}
\bibfield{author}{\bibinfo{person}{Claudiu Cob{\^{a}}rzan},
  \bibinfo{person}{Klaus Schoeffmann}, \bibinfo{person}{Werner Bailer},
  \bibinfo{person}{Wolfgang H{\"{u}}rst}, \bibinfo{person}{Adam Bla\v{z}ek},
  \bibinfo{person}{Jakub Loko\v{c}}, \bibinfo{person}{Stefanos Vrochidis},
  \bibinfo{person}{Kai~Uwe Barthel}, {and} \bibinfo{person}{Luca Rossetto}.}
  \bibinfo{year}{2017}\natexlab{}.
\newblock \showarticletitle{Interactive video search tools: a detailed analysis
  of the video browser showdown 2015}.
\newblock \bibinfo{journal}{\emph{Multimedia Tools Appl.}}
  \bibinfo{volume}{76}, \bibinfo{number}{4} (\bibinfo{year}{2017}),
  \bibinfo{pages}{5539--5571}.
\newblock
\urldef\tempurl%
\url{https://doi.org/10.1007/s11042-016-3661-2}
\showDOI{\tempurl}


\bibitem[\protect\citeauthoryear{Gygli}{Gygli}{2018}]%
        {Gygli18}
\bibfield{author}{\bibinfo{person}{Michael Gygli}.}
  \bibinfo{year}{2018}\natexlab{}.
\newblock \showarticletitle{Ridiculously Fast Shot Boundary Detection with
  Fully Convolutional Neural Networks}. In \bibinfo{booktitle}{\emph{2018
  International Conference on Content-Based Multimedia Indexing, {CBMI} 2018,
  La Rochelle, France, September 4-6, 2018}}. \bibinfo{pages}{1--4}.
\newblock
\urldef\tempurl%
\url{https://doi.org/10.1109/CBMI.2018.8516556}
\showDOI{\tempurl}


\bibitem[\protect\citeauthoryear{Hassanien, Elgharib, Selim, Hefeeda, and
  Matusik}{Hassanien et~al\mbox{.}}{2017}]%
        {Hassanien17}
\bibfield{author}{\bibinfo{person}{Ahmed Hassanien},
  \bibinfo{person}{Mohamed~A. Elgharib}, \bibinfo{person}{Ahmed Selim},
  \bibinfo{person}{Mohamed Hefeeda}, {and} \bibinfo{person}{Wojciech Matusik}.}
  \bibinfo{year}{2017}\natexlab{}.
\newblock \showarticletitle{Large-scale, Fast and Accurate Shot Boundary
  Detection through Spatio-temporal Convolutional Neural Networks}.
\newblock \bibinfo{journal}{\emph{CoRR}}  \bibinfo{volume}{abs/1705.03281}
  (\bibinfo{year}{2017}).
\newblock
\showeprint[arxiv]{1705.03281}
\urldef\tempurl%
\url{http://arxiv.org/abs/1705.03281}
\showURL{%
\tempurl}


\bibitem[\protect\citeauthoryear{{Huan}, {Xiuhuan}, and {Lilei}}{{Huan}
  et~al\mbox{.}}{2008}]%
        {Huan2008}
\bibfield{author}{\bibinfo{person}{Z. {Huan}}, \bibinfo{person}{L. {Xiuhuan}},
  {and} \bibinfo{person}{Y. {Lilei}}.} \bibinfo{year}{2008}\natexlab{}.
\newblock \showarticletitle{Shot Boundary Detection Based on Mutual Information
  and Canny Edge Detector}. In \bibinfo{booktitle}{\emph{2008 International
  Conference on Computer Science and Software Engineering}},
  Vol.~\bibinfo{volume}{2}. \bibinfo{pages}{1124--1128}.
\newblock
\urldef\tempurl%
\url{https://doi.org/10.1109/CSSE.2008.939}
\showDOI{\tempurl}


\bibitem[\protect\citeauthoryear{Kingma and Ba}{Kingma and Ba}{2014}]%
        {Adam14}
\bibfield{author}{\bibinfo{person}{Diederik~P. Kingma} {and}
  \bibinfo{person}{Jimmy Ba}.} \bibinfo{year}{2014}\natexlab{}.
\newblock \showarticletitle{Adam: {A} Method for Stochastic Optimization}.
\newblock \bibinfo{journal}{\emph{CoRR}}  \bibinfo{volume}{abs/1412.6980}
  (\bibinfo{year}{2014}).
\newblock
\showeprint[arxiv]{1412.6980}
\urldef\tempurl%
\url{http://arxiv.org/abs/1412.6980}
\showURL{%
\tempurl}


\bibitem[\protect\citeauthoryear{Loko\v{c}, Bailer, Schoeffmann, M{\"{u}}nzer,
  and Awad}{Loko\v{c} et~al\mbox{.}}{2018}]%
        {LokocBSMA18}
\bibfield{author}{\bibinfo{person}{Jakub Loko\v{c}}, \bibinfo{person}{Werner
  Bailer}, \bibinfo{person}{Klaus Schoeffmann}, \bibinfo{person}{Bernd
  M{\"{u}}nzer}, {and} \bibinfo{person}{George Awad}.}
  \bibinfo{year}{2018}\natexlab{}.
\newblock \showarticletitle{On Influential Trends in Interactive Video
  Retrieval: Video Browser Showdown 2015-2017}.
\newblock \bibinfo{journal}{\emph{{IEEE} Trans. Multimedia}}
  \bibinfo{volume}{20}, \bibinfo{number}{12} (\bibinfo{year}{2018}),
  \bibinfo{pages}{3361--3376}.
\newblock
\urldef\tempurl%
\url{https://doi.org/10.1109/TMM.2018.2830110}
\showDOI{\tempurl}


\bibitem[\protect\citeauthoryear{Loko\v{c}, Koval\v{c}\'{\i}k, M\"{u}nzer,
  Sch\"{o}ffmann, Bailer, Gasser, Vrochidis, Nguyen, Rujikietgumjorn, and
  Barthel}{Loko\v{c} et~al\mbox{.}}{2019}]%
        {Lokoc2019}
\bibfield{author}{\bibinfo{person}{Jakub Loko\v{c}}, \bibinfo{person}{Gregor
  Koval\v{c}\'{\i}k}, \bibinfo{person}{Bernd M\"{u}nzer},
  \bibinfo{person}{Klaus Sch\"{o}ffmann}, \bibinfo{person}{Werner Bailer},
  \bibinfo{person}{Ralph Gasser}, \bibinfo{person}{Stefanos Vrochidis},
  \bibinfo{person}{Phuong~Anh Nguyen}, \bibinfo{person}{Sitapa
  Rujikietgumjorn}, {and} \bibinfo{person}{Kai~Uwe Barthel}.}
  \bibinfo{year}{2019}\natexlab{}.
\newblock \showarticletitle{Interactive Search or Sequential Browsing? A
  Detailed Analysis of the Video Browser Showdown 2018}.
\newblock \bibinfo{journal}{\emph{ACM Trans. Multimedia Comput. Commun. Appl.}}
  \bibinfo{volume}{15}, \bibinfo{number}{1}, Article \bibinfo{articleno}{29}
  (\bibinfo{date}{Feb.} \bibinfo{year}{2019}), \bibinfo{numpages}{18}~pages.
\newblock
\showISSN{1551-6857}
\urldef\tempurl%
\url{https://doi.org/10.1145/3295663}
\showDOI{\tempurl}


\bibitem[\protect\citeauthoryear{Pass, Zabih, and Miller}{Pass
  et~al\mbox{.}}{1996}]%
        {Pass1997}
\bibfield{author}{\bibinfo{person}{Greg Pass}, \bibinfo{person}{Ramin Zabih},
  {and} \bibinfo{person}{Justin Miller}.} \bibinfo{year}{1996}\natexlab{}.
\newblock \showarticletitle{Comparing Images Using Color Coherence Vectors}. In
  \bibinfo{booktitle}{\emph{Proceedings of the Fourth ACM International
  Conference on Multimedia}} \emph{(\bibinfo{series}{MULTIMEDIA '96})}.
  \bibinfo{publisher}{ACM}, \bibinfo{address}{New York, NY, USA},
  \bibinfo{pages}{65--73}.
\newblock
\showISBNx{0-89791-871-1}
\urldef\tempurl%
\url{https://doi.org/10.1145/244130.244148}
\showDOI{\tempurl}


\bibitem[\protect\citeauthoryear{Shao, Qu, and Cui}{Shao et~al\mbox{.}}{2015}]%
        {shao15}
\bibfield{author}{\bibinfo{person}{Hong Shao}, \bibinfo{person}{Yang Qu}, {and}
  \bibinfo{person}{Wencheng Cui}.} \bibinfo{year}{2015}\natexlab{}.
\newblock \showarticletitle{Shot boundary detection algorithm based on HSV
  histogram and HOG feature}.
\newblock \bibinfo{journal}{\emph{5th International Conference on Advanced
  Engineering Materials and Technology}} (\bibinfo{year}{2015}).
\newblock


\bibitem[\protect\citeauthoryear{Tang, Feng, Kuang, Chen, and Zhang}{Tang
  et~al\mbox{.}}{2018}]%
        {Tang2018}
\bibfield{author}{\bibinfo{person}{Shitao Tang}, \bibinfo{person}{Litong Feng},
  \bibinfo{person}{Zhanghui Kuang}, \bibinfo{person}{Yimin Chen}, {and}
  \bibinfo{person}{Wei Zhang}.} \bibinfo{year}{2018}\natexlab{}.
\newblock \showarticletitle{Fast Video Shot Transition Localization with Deep
  Structured Models}.
\newblock \bibinfo{journal}{\emph{CoRR}}  \bibinfo{volume}{abs/1808.04234}
  (\bibinfo{year}{2018}).
\newblock
\showeprint[arxiv]{1808.04234}
\urldef\tempurl%
\url{http://arxiv.org/abs/1808.04234}
\showURL{%
\tempurl}


\bibitem[\protect\citeauthoryear{van~den Oord, Dieleman, Zen, Simonyan,
  Vinyals, Graves, Kalchbrenner, Senior, and Kavukcuoglu}{van~den Oord
  et~al\mbox{.}}{2016}]%
        {oord16}
\bibfield{author}{\bibinfo{person}{A{\"{a}}ron van~den Oord},
  \bibinfo{person}{Sander Dieleman}, \bibinfo{person}{Heiga Zen},
  \bibinfo{person}{Karen Simonyan}, \bibinfo{person}{Oriol Vinyals},
  \bibinfo{person}{Alex Graves}, \bibinfo{person}{Nal Kalchbrenner},
  \bibinfo{person}{Andrew~W. Senior}, {and} \bibinfo{person}{Koray
  Kavukcuoglu}.} \bibinfo{year}{2016}\natexlab{}.
\newblock \showarticletitle{WaveNet: {A} Generative Model for Raw Audio}.
\newblock \bibinfo{journal}{\emph{CoRR}}  \bibinfo{volume}{abs/1609.03499}
  (\bibinfo{year}{2016}).
\newblock
\showeprint[arxiv]{1609.03499}
\urldef\tempurl%
\url{http://arxiv.org/abs/1609.03499}
\showURL{%
\tempurl}


\bibitem[\protect\citeauthoryear{Zhang, Kankanhalli, and Smoliar}{Zhang
  et~al\mbox{.}}{1993}]%
        {zhang93}
\bibfield{author}{\bibinfo{person}{HongJiang Zhang}, \bibinfo{person}{Atreyi
  Kankanhalli}, {and} \bibinfo{person}{Stephen~W. Smoliar}.}
  \bibinfo{year}{1993}\natexlab{}.
\newblock \showarticletitle{Automatic partitioning of full-motion video}.
\newblock \bibinfo{journal}{\emph{Multimedia Systems}} \bibinfo{volume}{1},
  \bibinfo{number}{1} (\bibinfo{date}{01 Jan} \bibinfo{year}{1993}),
  \bibinfo{pages}{10--28}.
\newblock
\showISSN{1432-1882}
\urldef\tempurl%
\url{https://doi.org/10.1007/BF01210504}
\showDOI{\tempurl}


\end{thebibliography}

\end{document}